%% file: paper_1.tex
\documentclass[10pt,twocolumn,letterpaper]{article}

\usepackage[pagenumbers]{wacv} 

\usepackage{graphicx}
\usepackage{amsmath}
\usepackage{amssymb}
\usepackage{booktabs}
\usepackage{xspace}
\usepackage{multirow,longtable}
\newcommand{\name}[0]{{\sc EfficientMorph}\@\xspace}

\usepackage[pagebackref,breaklinks,colorlinks]{hyperref}
\usepackage{rotating}

\usepackage[capitalize]{cleveref}
\crefname{section}{Sec.}{Secs.}
\Crefname{section}{Section}{Sections}
\Crefname{table}{Table}{Tables}
\crefname{table}{Tab.}{Tabs.}

\def\wacvPaperID{2700} 
\def\confName{WACV}
\def\confYear{2025}

\begin{document}

\title{\name: Parameter-Efficient Transformer-Based Architecture for 3D Image Registration}

\author{Abu Zahid Bin Aziz $^{*}$ 
\and
Mokshagna Sai Teja Karanam$^{*}$ 
\and
Tushar Kataria 
\and
Shireen Y. Elhabian$^{**}$ \\
{\tt\small \{zahid.aziz,mkaranam,tushar.kataria,shireen\}@sci.utah.edu} \\
Scientific Computing and Imaging Institute \& Kahlert School of Computing \\
University of Utah, Salt Lake City, UT, USA \\
$^{*}${\textbf{Equal Contribution}}, $^{**}${\textbf{Corresponding Author}}
}

\maketitle

\input{Abstract}
\input{Introduction}

\input{RelatedWorks}

\input{Methods}

\input{Results}

\input{conclusion}

{\small
\bibliographystyle{ieee_fullname}
\bibliography{paper_1}
}
\newpage
\input{Supplementary}

\end{document}

%% file: Abstract.tex
\begin{abstract}
Transformers have emerged as the state-of-the-art architecture in medical image registration, outperforming convolutional neural networks (CNNs) by addressing their limited receptive fields and overcoming gradient instability in deeper models. 
Despite their success, transformer-based models require substantial resources for training, including data, memory, and computational power, which may restrict their applicability for end users with limited resources.
In particular, existing transformer-based 3D image registration architectures face two critical gaps that challenge their efficiency and effectiveness. 
Firstly, although window-based attention mechanisms reduce the quadratic complexity of full attention by focusing on local regions, they often struggle to effectively integrate both local and global information.
Secondly, the granularity of tokenization, a crucial factor in registration accuracy, presents a performance trade-off: smaller voxel-size tokens enhance detail capture but come with increased computational complexity, higher memory usage, and a greater risk of overfitting.
We present \name, a transformer-based architecture for unsupervised 3D image registration that balances local and global attention in 3D volumes through a plane-based attention mechanism and employs a Hi-Res tokenization strategy with merging operations, thus capturing finer details without compromising computational efficiency.
Notably, \name sets a new benchmark for performance on the OASIS dataset with $\sim$16-27$\times$ fewer parameters. \href{https://github.com/MedVIC-Lab/Efficient_Morph_Registration}{https://github.com/MedVIC-Lab/Efficient\_Morph\_Registration}
\end{abstract}

%% file: Introduction.tex
\section{Introduction}
3D image registration \cite{viergever2016survey,hering2022learn2reg} is a critical task for various medical imaging applications in fields such as image-guided surgery \cite{alam2018medical}, radiation therapy planning \cite{rigaud2019deformable}, image fusion for multimodality imaging \cite{huang2022reconet}, and quality enhancement \cite{azam2021multimodal}. 
Registration involves determining the spatial alignment between two volumes, typically referred to as the \textit{fixed} and \textit{moving} images,  by identifying correspondences between similar structures or features and their relative spatial positions.
Conventional approaches such as ANTs \cite{avants2009advanced}, Elastix \cite{klein2009elastix}, and NiftiReg \cite{modat2010fast} employ optimization-based frameworks \cite{wells1996multi,heinrich2012mind,avants2008symmetric,glocker2011deformable}. 
This iterative search for the optimal transformation makes these methods inherently slow, especially when dealing with large datasets or high-resolution images \cite{li2008accuracy,mujat2023non,kataria2023automating}. 
To address these challenges, there is increasing interest in adopting learning-based approaches. In particular, deep learning methods offer significantly faster inference times and currently achieve state-of-the-art performance in 3D image registration \cite{chen2022transmorph,jia2023fourier,balakrishnan2019voxelmorph}.
\begin{figure}[!tb]
    \centering
    \includegraphics[scale=0.18]{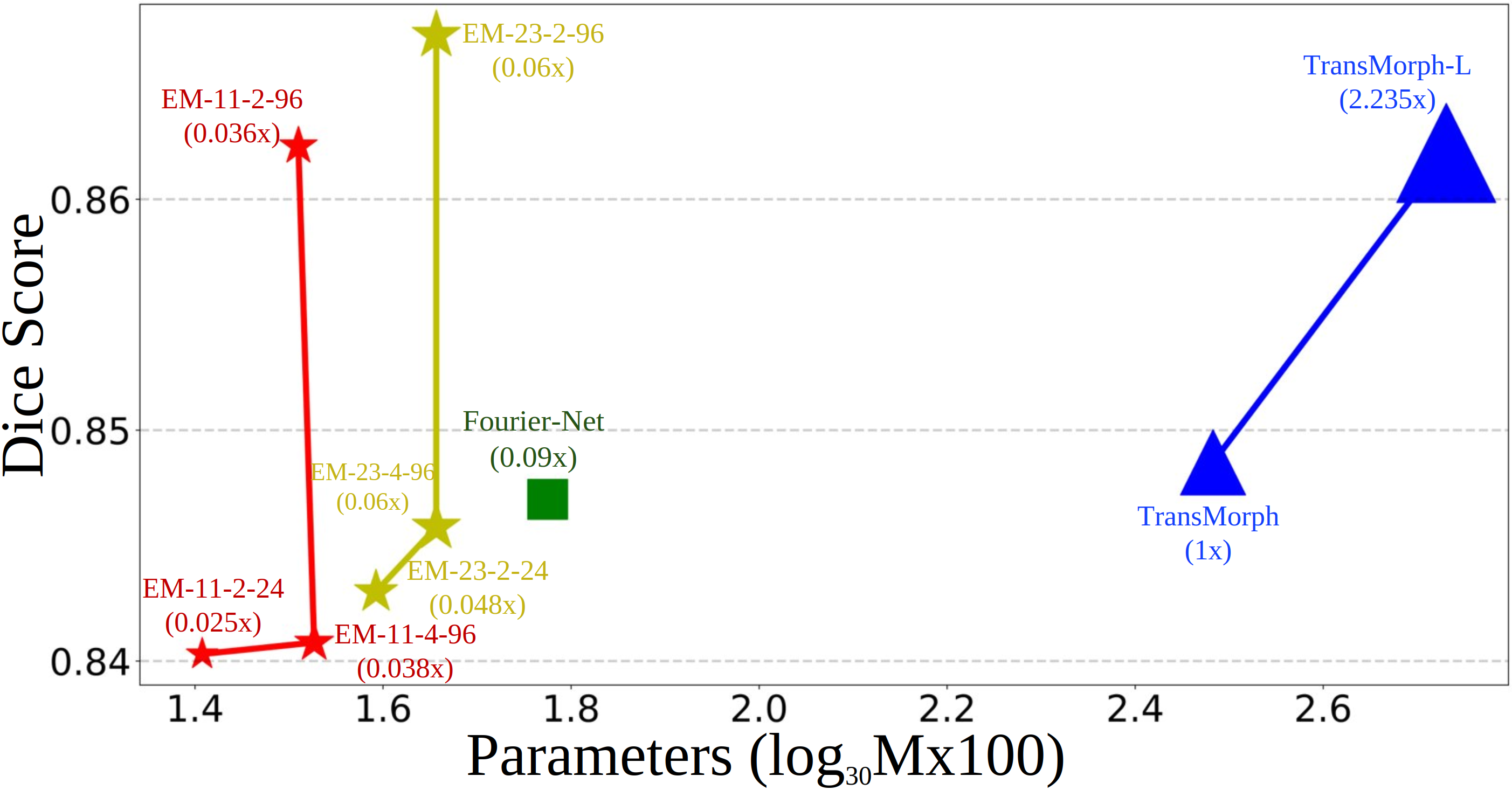}
    \caption{\textbf{Parameter Count Comparisons with performance on OASIS Dataset.} The proposed variants are formatted as EfficientMorph-11-stride-$C$ and EfficientMorph-23-stride-$C$. Comparison of parameter count in millions(M) and Dice scores between the proposed variants and baselines. }
    \label{fig:Parameter_Count}
    \vspace{-1em}
\end{figure}

Learning-based registration methods predominantly rely on convolutional architectures (e.g., \cite{balakrishnan2019voxelmorph,kim2021cyclemorph,jia2023fourier,mok2020large,10052953}), using U-Net-based architectures to generate the deformation fields.
However, the effectiveness of convolutional layers for registration tasks can be compromised due to their limited receptive fields that hinder capturing global context \cite{chen2022transmorph} and their increased susceptibility to vanishing gradients as network depth grows to enhance learning capacity \cite{hanin2018neural}.
Since the advent of Vision Transformers \cite{vaswani2017attention}, transformer-based architectures have shown superior performance across various tasks, such as classification, segmentation, and registration \cite{vaswani2017attention,xie2021segformer,chen2022transmorph,dosovitskiy2020image}, thanks to their long-range modeling capabilities. 
In particular, they offer promising mitigations to CNN limitations. Specifically, transformers leverage global contextual information through self-attention mechanisms and provide more stable gradient flow across layers via techniques such as layer normalization and skip connections that are integral to transformers' design \cite{vaswani2017attention,chen2022transmorph,dosovitskiy2020image}.
Despite their success, transformers' advantages come at the expense of a significant increase in parameter count, requiring approximately 10 to 20 times more parameters than convolutional counterparts \cite{chen2022transmorph,ghahremani2024h}, making them impractical for deployment to end-user applications.

Specifically, existing transformer-based registration methods, including TransMorph \cite{chen2022transmorph}, the current state-of-the-art transformer-based model for medical image registration, encounter \textit{two} main significant limitations that compromise its efficiency and overall performance.
\textit{Firstly}, windowed attention approaches (e.g., the Swin transformer \cite{liu2021swin} backbone used in TransMorph \cite{chen2022transmorph}) optimize computational efficiency through local attention and shifted windows, enhancing interactions between adjacent windows. However, this limits global context capture, particularly in shallow layers, due to within-window constraints(masks for calculating attention) compared to methods that interact globally. 
\textit{Secondly}, the pixel granularity of tokenization plays a crucial role in registration accuracy. To fit within available GPU memory, tokenization is applied to downsampled volumes, with each dimension reduced by 4 to 8. Increasing the token resolution within the same volume can capture finer details, but it also escalates computational and memory demands due to the corresponding rise in the number of tokens. \cite{chen2022transmorph,vaswani2017attention,keles2023computational}. Existing multi-resolution architectures, such as GradICON \cite{tian2023gradicon}, HRNet \cite{wang2020deep}, and HRFormer \cite{yuan2021hrformer}, leverage features at various resolutions to enhance performance. However, these approaches necessitate the simultaneous training of multiple models to optimize performance at each resolution level. These training methods significantly increase complexity and computational cost, leading to a substantial rise in the number of parameters. As a result, the models become more resource-intensive and challenging to deploy in environments with limited resources.

In this paper, we propose \name, a novel transformer-based framework for unsupervised 3D image registration that addresses the aforementioned challenges. We introduce a \textit{plane attention} mechanism inspired by 3D anatomical views (coronal, sagittal, and axial) to enhance the balance between local and global feature recognition \cite{ho2019axial}. 
We propose Hi-Resolution tokenization to capture finer image details. To further reduce model complexity within the encoded representation, we introduce a method for merging neighboring tokens in a high-resolution feature space, thereby decreasing the computational load of self-attention calculations. By integrating Hi-Res tokenization, \name becomes a highly parameter-efficient registration architecture (see Figure \ref{fig:Parameter_Count}A). Additionally, we introduce a multi-resolution \name, which concatenates latent features from different resolutions to produce more precise deformation fields. This approach leverages multi-resolution data without needing to train separate models.

\vspace{0.05in}
\noindent The main contributions of this paper are:
\vspace{-0.5em}
\begin{itemize}
\setlength\itemsep{-0.2em}
    \item A novel attention module for 3D registration that focuses attention across the coronal ($xy$), sagittal ($yz$), or Axial ($zx$) planes within a single transformer block.
    \item A Hi-Resolution tokenization mechanism to encode high-resolution voxel features without increasing computation complexity.
    \item Proposed Multi-Resolution \name leverages the concatenation of multi-resolution latent space features to enhance model performance.
    \item A new parameter-efficient architecture achieves performance within ±0.05 Dice score of existing methods, surpassing state-of-the-art on 2 out of 3 datasets (single and multi-modal Registration), with 16-27x fewer parameters (Figure \ref{fig:Parameter_Count}) and ~5x faster convergence. Comprehensive ablation studies on regularization losses, attention mechanisms, and key design choices are also presented.
\end{itemize}

%% file: RelatedWorks.tex
\section{Related Works}

\noindent\textbf{3D Volume Registration.}
Learning-based approaches for 3D image registration can generally be divided into two main categories: supervised and unsupervised.
\textit{Supervised methods} \cite{sokooti2017nonrigid,yang2016fast,sokooti20193d} require estimates of deformation fields derived from traditional optimization-based approaches, the acquisition of which can be prohibitively costly for large datasets. 
Moreover, the efficacy of supervised approaches is contingent upon the availability of high-quality deformation fields for supervised training, with their performance capped by the accuracy of the method used to obtain these fields. 
In contrast, \textit{unsupervised methods} do not require deformation fields and use image similarity as a self-supervised signal to train a registration network.
Most unsupervised 3D registration methods  \cite{balakrishnan2019voxelmorph,chen2022transmorph,jia2023fourier,mok2020large,meng2022non} are trained to produce a 3D deformation field that is then used to transform (or warp) the moving image. Loss (L1 or L2) between the warped moving image and the fixed image is used to train the network. With sufficient data and training time, the model learns to produce realistic deformation fields that outperform optimization-based methods in accuracy and inference speed \cite{chen2022transmorph,jia2023fourier}. Additionally, unsupervised methods incorporate regularization losses to promote spatial smoothness in the deformation field, often employing techniques such as bending energy \cite{rueckert1999nonrigid,johnson2002consistent}, total variation minimization \cite{vishnevskiy2016isotropic}, and consistency penalties \cite{tian2024unigradicon,tian2023gradicon,zhang2018inverse}, among others. To enhance the accuracy of registration, segmentations of the underlying anatomies are incorporated as regularization losses \cite{chen2022transmorph,jia2023fourier}. However, this approach makes the registration problem fully or semi-supervised due to the requirement for manual segmentation. Ideally, unsupervised registration methods should perform effectively without the need for additional supervision. We present a parameter-efficient registration architecture that outperforms state-of-the-art models on three public datasets while significantly reducing the number of parameters. Our proposed model not only performs well in unsupervised 3D volume registration but can also leverage available segmentation data to outperform state-of-the-art models, all while maintaining a lower parameter count.

\textbf{Efficient Transformer Attention Architectures.} 
As deep learning models continue to grow in size each year \cite{woo2023convnext,vorontsov2023virchow,achiam2023gpt,dubey2024llama}, deploying them on end-user devices such as mobile phones or desktops becomes increasingly impractical. Users often need access to a server API for model inference or a local desktop with substantial computing power to run these models. These requirements restrict the applicability of many deep learning models, particularly in medical applications where data privacy is paramount \cite{mireshghallah2020privacy,yu2021large,jin2019review}. In many cases, patient data cannot be transferred to a server, requiring computations to be performed locally to comply with HIPAA guidelines. As a result, developing efficient architectures that preserve the accuracy of large models while being deployable on end-user devices is both essential and highly relevant. Examples of such architectures and methods proposed for efficient deep learning models include EfficientNet \cite{tan2019efficientnet}, MobileNet \cite{howard2017mobilenets}, LLM-pruner \cite{ma2023llm}, GPTQ \cite{frantar2022gptq}, and Mobillama \cite{thawakar2024mobillama} \cite{wang2024computation,wang2023closer,bartoldson2023compute,kaddour2024no}.

Applying transformer self-attention to high-resolution medical images presents substantial computational challenges due to its quadratic complexity with respect to input size \cite{keles2023computational,vaswani2017attention,zhao2020exploring}, making it difficult and resource-intensive to scale to large datasets and model sizes. As a result, these models often cannot be deployed in end-user applications in hospitals and clinics. Various strategies have been developed to address the computational challenges of applying transformer self-attention to high-resolution images. 
An effective strategy involves optimizing the attention matrix through techniques such as approximations—like Linformer \cite{wang2020linformer}, Memory efficient attention \cite{rabe2021self}, and sparse attention \cite{child2019generating,roy2021efficient}—or by limiting exact attention to localized windows, as seen in models like the SWIN Transformer \cite{liu2021swin}. Moreover, efficiency can be significantly improved through GPU optimizations, as demonstrated by Flash Attention \cite{dao2022flashattention,dao2023flashattention}. An alternative strategy is to stack multiple sparse attention layers with restricted contexts, which allows overlapping layers to achieve full-context modeling. For example, the Strided Sparse Transformer \cite{fan2022embracing} employs custom GPU kernels to implement block-sparse matrix multiplications, enhancing computational efficiency. Similarly, the Axial Transformer \cite{ho2019axial} maintains full conditioning contexts by processing both masked and unmasked tokens during each decomposition stage. In contrast, our proposed module is specifically designed for 3D medical volumes, introducing a novel 3D \textit{plane-based} attention mechanism that selectively operates on a subset of planes (\textit{axial, sagittal, coronal}) within each decomposition block. This approach allows for the creation of models with significantly fewer parameters. Furthermore, by utilizing high-resolution voxel tokens, our model matches and surpasses the performance of state-of-the-art models.

%% file: Methods.tex
\begin{figure*}[!htb]
    \centering
    \includegraphics[scale=0.23]{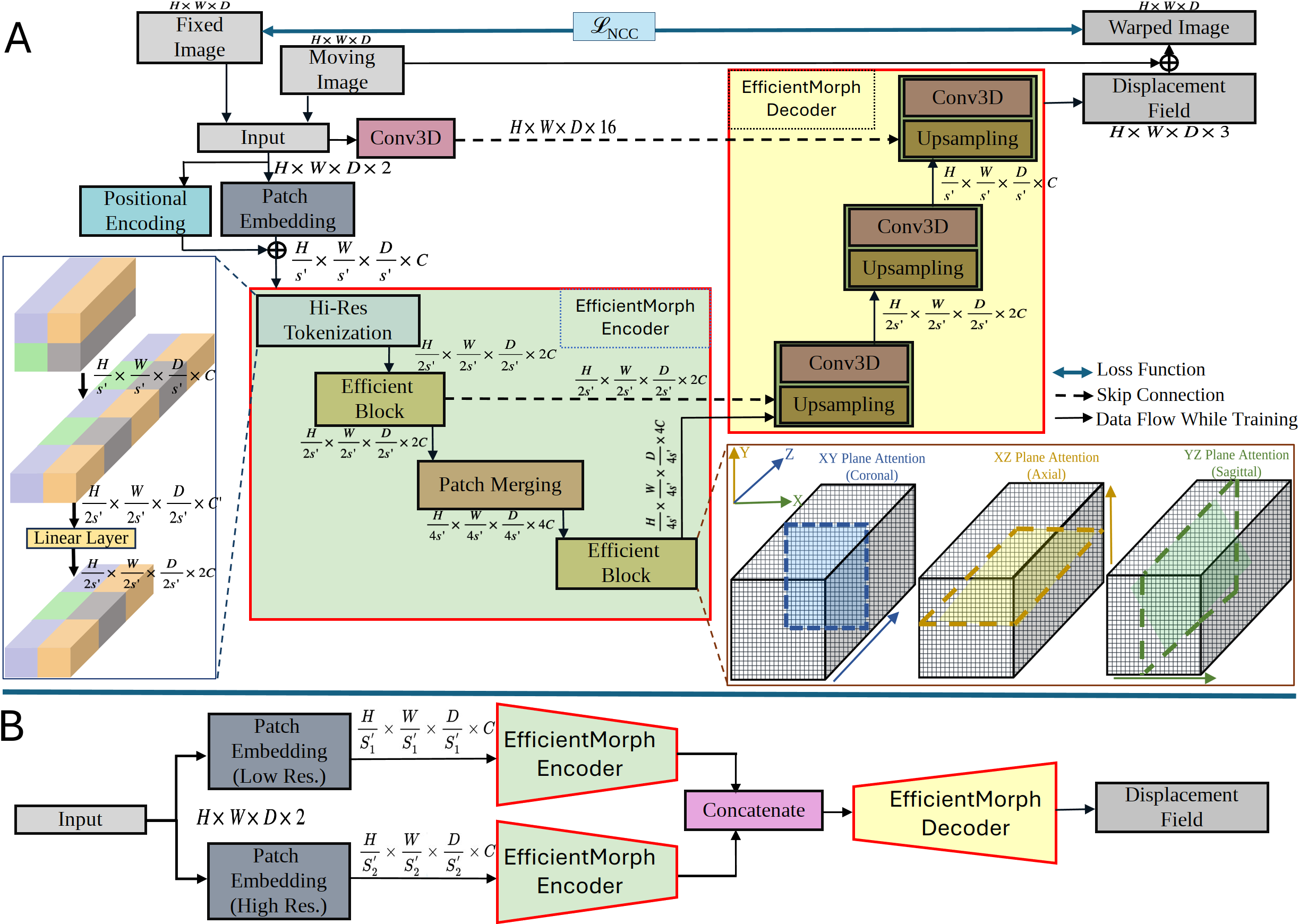}
    \caption{\textbf{\name Architecture.} 
    (A) \name utilizes utilizes plane attention mechanism on the whole volume as shown in \textit{Efficient Transformer Block}.
     We use different numbers and types of plane attentions ($xy, yz$, or $zx$ planes) for each block in the transformer backbone (Table \ref{tab:EfficientMorph_Variations}). Hi-Res Tokenization is shown in the left end of the figure. (B) Shows the architectural modification for multi-resolution variant where $S^{'}_{1} > S^{'}_{2}$.}
    \label{fig:architecture}
    \vspace{-1.0em}
\end{figure*}
\section{Methods}

Given a 3D volume represented as $\mathbf{A} \in \mathbb{\mathbb{R}}^{H \times W \times D}$, where $H$, $W$, and $D$ denote the height, width, and depth dimensions, respectively. If each voxel is mapped into the latent space, the number of tokens for the volume would be \textit{WHD}, which may exceeds the memory capacity of a GPU for a single sample. Therefore, strided convolutions are used in the patch embedding layer (with stride $s$) to project the voxels in $\mathbf{A}$ into a high-dimensional feature space, resulting in $\mathbf{A}' \in \mathbb{R}^{H' \times W' \times D' \times C}$, where $C$ is the embedding dimension, $(H', W', D')=(\frac{H}{s},\frac{W}{s},\frac{D}{s})$. The resulting feature space is tokenized to train the downstream transformer layers. In the following sections, we detail the Hi-Res tokenization process, and the plane attention mechanism of the proposed \name, as illustrated in Figure \ref{fig:architecture}A. We have provided a comprehensive explanation of the overall training process in Sections \ref{supp:additional_traiing_details} and \ref{supp:additional_multires_details} of the supplementary materials.

\subsection{Hi-Res Tokenization} \label{sec:appx:hi-res-tokenization}

For a fixed embedding dimension $C$, using each voxel of a 3D volume of size $1$x$1$x$1$ for tokenization would create $N$ tokens, where $N=H\times W\times D$. Voxel-level tokenization results in attention matrices of more than a trillion parameters with a complexity of $\mathcal{O}(N^2)$. Transformer architectures for images often rely on $s$-strided convolutions (e.g., $s = 4$ \cite{chen2022transmorph}) for volume tokenization and patch embedding, trading off computational complexity, which is now $\mathcal{O}\left(\left(\frac{N}{s^3}\right)^2\right)$, at the cost of detailed features. However, fine-grained spatial information is critical for medical segmentation and registration tasks, which may be lost due to strided convolutions. 

We propose \textit{Hi-Resolution tokenization} strategy that uses a smaller stride ($s'<4$) within the embedding layer, thereby utilizing better spatial information available in the volume. However, this increases the number of tokens increasing the computational complexity. To reduce the computation complexity, we propose a token merging operation. These high-resolution tokens are positionally encoded and merged by grouping and concatenating the features of adjacent non-overlapping $d \times d \times d$ voxel token blocks (along the embedding dimension), resulting in $N'=\frac{H}{d}\times\frac{W}{d}\times\frac{D}{d}$ tokens with an embedding dimension of $C'= C \times {d}^3$. Then, $C'$ is projected into a linear layer to attain a reduced embedding dimension of $C\times d$, as shown in Hi-Res tokenization block in Figure \ref{fig:architecture}A.  This approach to tokenization enables us to use high-resolution features and reduces the overall complexity of the model. Refer to Figure \ref{fig:architecture}A for pictorial depiction of the Hi-Res Tokenization process.

\subsection{Plane Attention Mechanism}
 Despite using Hi-Res tokenization, the number of tokens generated from each volume remains high. Running full attention on these tokens, while feasible, demands considerable computational resources. We introduce a novel attention framework called \textit{plane attention} to address this challenge. Instead of performing full 3D attention on all tokens, this method utilizes attention along coronal ($xy$), sagittal ($yz$), or axial ($zx$) planes, as shown in Figure \ref{fig:architecture}A. Although attention confines focus to a specific plane, \name achieves volume attention by sequentially employing different attention combinations $xy$ followed by $yz$ or $zx$, thus covering all plane directions.

\begin{equation}
    \text{Attn}(\mathbf{A}'_{dim}) = \text{softmax}\left(\frac{\mathbf{Q}_{dim}\mathbf{K}_{dim}^T}{\sqrt{d_k}}\right)\mathbf{V}_{dim}
\end{equation}
Here, $dim\in\{xy, yz, zx\}$, $\mathbf{A}'_{dim}$ can be represented as $\mathbf{A}'_{xy} \in \mathbb{\mathbf{R}}^{H' \times W' \times C}$, $\mathbf{A}'_{yz} \in \mathbb{\mathbf{R}}^{W' \times D' \times C}$ and $\mathbf{A}'_{zx} \in \mathbb{\mathbf{R}}^{ D' \times H' \times C}$ for $xy$, $yz$, and $zx$ planes, respectively. 

By decomposing the 3D attention into 2D plane attention, the proposed attention mechanism significantly reduces the parameter count while preserving the ability to capture essential volumetric features necessary for registration. Figure \ref{fig:architecture}A shows different plane attention blocks across the efficient transformer block.

\subsection{Multi-Resolution \name}
HiRes Tokenization effectively harnesses high-resolution spatial information early in the network but limits all tokens to a single resolution. However, prior work \cite{tian2023gradicon,tian2024unigradicon,yuan2021hrformer,wang2020deep} has demonstrated that multi-resolution features significantly improve registration accuracy. Therefore, we propose a multi-resolution variant for \name. 

Multi-resolution \name processes the input image along two distinct paths, each with its patch embedding block tailored to tokenize patches of different sizes ($S_1^{'}$ and $S_2^{'}$, where $S_1^{'} > S_2^{'}$). This approach captures patches at multiple resolutions, enhancing the richness of the feature representation. Instead of training the entire architecture simultaneously, a phased training strategy is adopted. In this process, patches pass through two stages of Efficient Transformer blocks, as illustrated in Figure \ref{fig:architecture}B, where their latent dimensions are merged. The merged representation is then fed into the decoder, ensuring that the output integrates comprehensive information from both resolution levels. This methodology not only improves the model's ability to capture fine details but also enhances computational efficiency during training.

%% file: Results.tex
\section{Results and Discussion}
\subsection{Datasets and Implementation Details}

\textbf{OASIS Brain MRI.} We evaluated \name on the publicly available dataset OASIS \cite{10.1162/jocn.2007.19.9.1498},  obtained from the Learn2Reg challenge \cite{hering2022learn2reg} for inter-patient registration and pre-processed from \cite{hoopes2021hypermorph}. It has a total of 451 brain T2 MRI images. Among these, 394, 19, and 38 scans are used for training, validation, and testing, respectively. 

\textbf{ReMIND2Reg.}  
This dataset aims to register 3D iUS images with either ceT1 or T2 MRI images to account for brain shift during tumor resection, requiring models to handle large deformations and missing data scenarios. The dataset is divided into 155 image pairs for training, 10 image pairs for validation, and 40 for testing. 

\textbf{Atlas-to-Patient Brain MRI (IXI).} We additionally evaluated the proposed model on IXI dataset that contains 600 MRI scans.
Among these, 576 T1-weighted brain MRI images were employed as moving images, while the fixed image for this task was an atlas brain MRI \cite{kim2021cyclemorph}. The dataset was partitioned into training, validation, and test sets, comprising 403, 58, and 115 volumes, respectively. For more details on datasets, refer to supplementary section \ref{supp:additional_dataset_details}

\vspace{0.05in}
\noindent \textbf{Implementation Details.}
\name was trained on NVIDIA A100 GPUs with 40GB RAM and a batch size of 1. We used the same splits for both datasets as the existing works \cite{chen2022transmorph,jia2023fourier}. 
We limited training epochs to 100 to prioritize parameter efficiency and quick convergence within resource limits. 
We used the Adam optimizer with a learning rate of $5e^{-4}$ for OASIS \& Remind2Reg and $3e^{-4}$ for IXI.  We used a cosine annealing scheduler for OASIS and stepLR for IXI. 
We evaluated different variants: EfficientMorph-11, which includes one plane attention transformer (\textit{xy}, \textit{yz}, or \textit{zx}) per efficient block as shown in Figure \ref{fig:architecture}, and EfficientMorph-23, which features two plane attention transformers in the first efficient block and three in the second. The specific plane attentions used in these variants are detailed in Table \ref{tab:EfficientMorph_Variations}. Note that no data augmentation was applied during training.

\begin{table}[!htb]
\centering
\caption{\textbf{EfficientMorph Variants.} \name-AB denotes a configuration with A plane attention transformers in the first efficient block and B plane attention transformers in the second efficient block.}
\label{tab:EfficientMorph_Variations}
\vspace{-0.5em}
\small
\setlength{\tabcolsep}{3pt} 
\scalebox{1}{
\begin{tabular}{c|c} 
 Variants  & Planes\\ \hline
\name-11  & (xy, yz)\\ \hline
\name-23  & (xy-yz, xy-yz-zx)
\end{tabular}}
\end{table}

\textbf{Loss function.} In the unsupervised registration setting, we utilized normalized cross-correlation with bending energy regularization, consistent with other registration frameworks in the literature \cite{chen2022transmorph,chen2023transmatch,jia2023fourier}. Let $I_{F}$ and $I_{M}$ be the fixed and moving image volumes and $S_{F}$ and $S_{M}$ be the associated anatomy segmentations (if available).
\begin{align*}
\mathcal{L_\text{UnSupReg}} = L_{NCC}(I_{F}, \text{Warp}(I_{M})) + \text{BendingEnergy}
\end{align*}
 To ensure a thorough comparison on the OASIS dataset, we also incorporated an additional segmentation loss (Dice coefficient) during training, aligning with the approaches used in other methods \cite{chen2022transmorph, jia2023fourier, balakrishnan2019voxelmorph}.
\begin{align*}
\mathcal{L_\text{OASIS}} =& L_{NCC}(I_{F}, \text{Warp}(I_{M})) + \text{BendingEnergy} \\ &+ \text{DiceLoss}(S_{F}, \text{Warp}(S_{M}))
\end{align*}

\paragraph{Comparisons Methods.} We compare \name with convolutional-based methods, including VoxelMorph \cite{balakrishnan2019voxelmorph} and Fourier-Net \cite{jia2023fourier}, as well as different Tranformer based methods such as TransMorph \cite{chen2022transmorph}, including TransMorph-Tiny, TransMorph, and TransMorph-L, TransMatch \cite{chen2023transmatch} and Vit-V-Net\cite{chen2021vit}. All methods were trained on the same GPU as previously mentioned, using their original implementations.

\textbf{Evaluation Metrics}. To evaluate the results on the OASIS and IXI datasets, we utilized the Dice score for anatomical segmentation (35 regions for OASIS and 29 for IXI) and computed the percentage of negative values of Jacobian determinant. For the ReMIND2Reg, we used the Learn2Reg \cite{hering2022learn2reg} leaderboard evaluation system, where the output deformation fields were submitted to obtain Target Registration Error (TRE) and percentage of negative values of Jacobian determinant of deformation.

\subsection{Experimental Results}

\textbf{OASIS Results.} The results on the OASIS dataset are shown in Table \ref{tab:OASIS:Results}. 
Among the variants, EfficientMorph-23 achieves the highest Dice score with just 2.8M parameters—16 times fewer than TransMorph and 8 times fewer than TransMatch—outperforming all compared baselines, including TransMorph-L, which has over 100M parameters. Despite having fewer parameters, EfficientMorph-11 delivers comparable performance to the other baselines. Both variants maintain consistently low percentage of negative values in Jacobian determinant of deformation, demonstrating that even with fewer parameters, \name learns a more robust representation of the underlying data, leading to superior registration performance. Table \ref{tab:OASIS:Results} further demonstrates the results when segmentation loss is added as an additional training loss for the registration network. While all models show improved accuracy with added segmentation supervision, \name still outperforms all others, achieving the highest average Dice score and the lowest Jacobian determinant score.

Supplementary Figure \ref{fig:OASIS_boxplot} presents a comparison of dice scores between the \name variants and the baseline across different brain MR substructures, highlighting significant improvements with our proposed models. Our models consistently achieve the highest test dice scores across all brain segments. Supplementary Figure \ref{fig:OASIS_qualitative} also provides qualitative results of the segmentations obtained after registration of three anatomies, along with their corresponding dice scores. The figure includes the best, median, and worst-performing cases for analysis. Notably, the worst-performing case, characterized by a fixed image with non-smooth boundaries, challenges all models in registration accuracy; however, our proposed model still outperforms others, achieving the highest Dice score for the case.

\begin{table*}[!tbh]

\caption{\textbf{OASIS Registration Results Using Single Resolution.} Average Dice Score Evaluated over 35 anatomies and percentage of negative values in Jacobian determinant of deformation are obtained for all test samples. \textit{ w/o Seg Loss} is the full unsupervised registration setting where only similarity and regularization loss between fixed and moving images are used for training. For \textit{Seg Loss} setting, segmentation loss between segmentation of fixed and moving image anatomies are also used for training. \textit{Param} are listed in Millions of parameters used for training the model. We can clearly see that \name performs on par or better than other models with fewer parameters.$^\text{*}$ indicates the performance numbers taken from TransMorph\cite{chen2022transmorph} and Fourier-Net\cite{jia2023fourier}.} \label{tab:OASIS:Results}
\centering
\setlength{\tabcolsep}{4pt}
\scalebox{1}{

\begin{tabular}{c||c|c|c||c|c||c|c}
& && &\multicolumn{2}{c||}{\bf w/o Seg Loss} &  \multicolumn{2}{c}{\bf with Seg Loss} \\
\hline
 \bf Methods & \bf stride &\bf C & \bf Param & \bf Dice $\uparrow$ & $|\mathbf{J}|< 0$\%  $\downarrow$ & \bf Dice $\uparrow$ & $|\mathbf{J}|< 0$\% 
 $\downarrow$ \\
\hline
VoxelMorph\cite{balakrishnan2019voxelmorph}& -  & - & 0.063 & 0.6783 $\pm$ 0.039& 2.981 $\pm$ 0.105 & 0.78 $\pm$ 0.024& 0.1304$\pm$0.011 \\
 Fourier-Net\cite{jia2023fourier} & - & -& 4.19    &  0.770 $\pm$ 0.021 & 0.031 $\pm$ 0.003 & 0.847$\pm$0.013$^\text{*}$ & -\\
 ViT-V-Net\cite{chen2021vit} & 8x8x8& 252 & 9.8 &0.3632$\pm$0.0072 & 0.0149 $\pm$ 0.0001 & 0.4659$\pm$0.0052 & 0.1272 $\pm$ 0.0145 \\
TransMatch \cite{chen2023transmatch} & 4x4x4& 96 & 26.39 & 0.4037 $\pm$ 0.055 &  0.1167 $\pm$ 0.0082 & 0.4612$\pm$0.0582 & 0.0546 $\pm$ 0.0011    \\

 \hline
 TransMorph-Tiny \cite{chen2022transmorph} & 4x4x4 & 6 & 0.24  & 0.441 $\pm$ 0.021 & 0.013 $\pm$ 0.001 & 0.801$\pm$0.056 & 0.081 $\pm$ 0.010\\
 TransMorph \cite{chen2022transmorph} & 4x4x4 &96 &46.5  & 0.801 $\pm$ 0.003 & 0.03 $\pm$ 0.002 &0.8458$\pm$0.0137 & 0.119 $\pm$ 0.019\\
 TransMorph-L \cite{chen2022transmorph} & 4x4x4 &128 & 108.11  & 0.804 $\pm$ 0.024 & \bf 0.009 $\pm$ 0.001 & 0.862$\pm$0.014$^\text{*}$ & 0.128 $\pm$ 0.021$^\text{*}$\\
\hline
EfficientMorph-23 & 4x4x4 & 96 & 2.8 & 0.796 $\pm$ 0.035 & 0.091 $\pm$ 0.0006  & 0.846 $\pm$ 0.013 & 0.125 $\pm$ 0.020 \\ 
 EfficientMorph-11 & 2x2x2 &  96  & \bf 1.8  & 0.803 $\pm$ 0.070 & \bf 0.011 $\pm$ 0.002 & \bf 0.8623$\pm$0.0133 & \bf 0.010 $\pm$ 0.001\\
 EfficientMorph-23 & 2x2x2 & 96  & \bf 2.8  & \bf 0.810  $\pm$ 0.062 & \bf 0.010 $\pm$ 0.001 & \bf 0.870 $\pm$ 0.016 & \bf 0.017 $\pm$ 0.001\\ 
\end{tabular} 
}
\end{table*}

\textbf{\name has a significantly low parameter count and coverges faster.} Figure \ref{fig:Parameter_Count} (Page 1) shows that compared to TransMorph \cite{chen2022transmorph}, \name proposed architectures have between 2.5-6\% of the total parameters, however with comparable and even better dice scores for EfficientMorph-23 variant. Similarly, when compared with Fourier-Net \cite{jia2023fourier}, Efficient morph EM-11-2-96 has $\frac{1}{3}$rd parameter count and with higher dice score. These results clearly show that \name achieves better registration accuracy than other models and a very low parameter count. Models trained with segmentation loss were used for this analysis, as using an extra loss doesn't have an impact on number of parameters. 

\begin{figure}[!htb]
    \centering
    \includegraphics[scale=0.17]{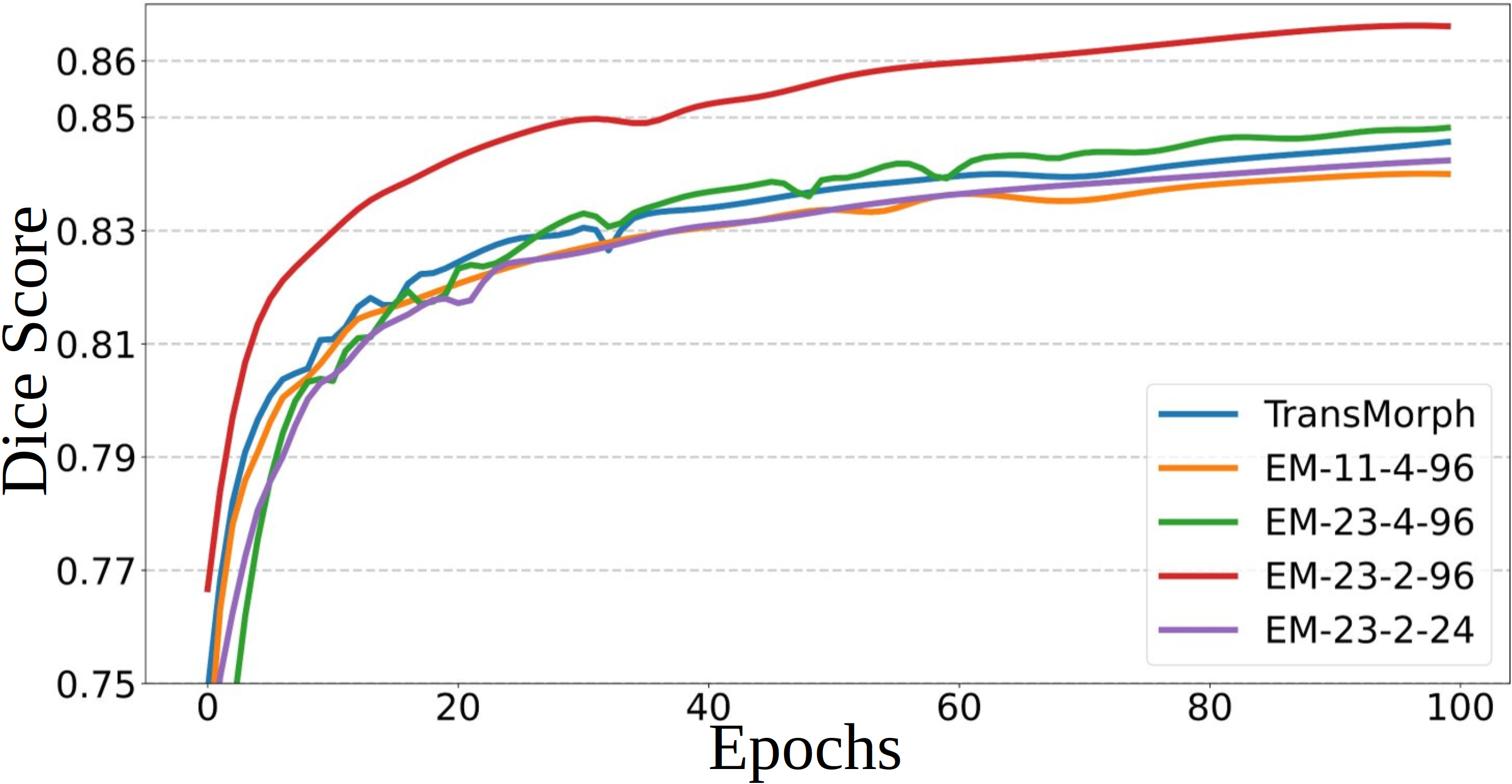}
    \caption{\textbf{Convergence Curves.} The proposed variants are formatted as EfficientMorph-11-stride-$C$ and EfficientMorph-23-stride-$C$. Dice score curves of EfficientMorph variants as a function of epochs. }\label{fig:Conversion_Curves}
\end{figure}

The convergence curves in Figure \ref{fig:Conversion_Curves} clearly show that TransMorph learns quickly in a few initial epochs but then slowly saturates to the final performance, whereas all EfficientMorph variants slowly and steadily converge to higher dice scores. EfficientMorph starts to outperform TransMorph by a significant margin as early as 10 epochs. This result clearly shows that efficient morph is not only parameter efficient but requires less compute for converging to a better solution. 
\begin{table}[!htb]
\caption{\textbf{Multiresolution Unsupervised Registration Results on OASIS.}} \label{tab:MR-OASIS:Results}
\centering
\small
\setlength{\tabcolsep}{2pt}
\scalebox{1}{
\begin{tabular}{p{0.24cm}|c||c|c|c}
 &\bf Methods & \bf stride & \bf Param & \bf Dice $\uparrow$  \\
 \hline
 \multirow{6}{=}{\begin{sideways}\bf Single Res \end{sideways}}
 &EM-11 &  ($2\times2\times2$)  & 1.8  & 0.803 $\pm$ 0.070  \\
 &EM-11 &  ($4\times4\times4$)  & 1.8  & 0.795 $\pm$ 0.071  \\
 &EM-11 &  ($8\times8\times8$)  & 1.8  & 0.765 $\pm$ 0.021  \\
 &EM-23 &  ($2\times2\times2$)  & 2.8  & 0.810  $\pm$ 0.062  \\
 &EM-23 &  ($4\times4\times4$)  & 2.8  & 0.796 $\pm$ 0.067  \\
 &EM-23 &  ($8\times8\times8$)  & 2.8  & 0.768 $\pm$ 0.026  \\
\hline
\multirow{6}{=}{\begin{sideways}\bf Multi Res \end{sideways}}
 &EM-11 &  ($2\times2\times2$),($4\times4\times4$)  & 6.8  & \bf 0.820 $\pm$ 0.041  \\
 &EM-11 & ($2\times2\times2$),($8\times8\times8$)  & 6.8   & \bf 0.821 $\pm$ 0.015 \\ 
 &EM-11 & ($4\times4\times4$),($8\times8\times8$)  & 6.8   &  0.812 $\pm$ 0.037 \\ 
 &EM-23 &  ($2\times2\times2$),($4\times4\times4$)  & 9.0  & 0.817 $\pm$ 0.023  \\
 &EM-23 & ($2\times2\times2$),($8\times8\times8$)  & 9.0   & \bf 0.818 $\pm$ 0.019 \\ 
 &EM-23 & ($4\times4\times4$),($8\times8\times8$)  & 9.0   & 0.811 $\pm$ 0.021 \\ 
\end{tabular} 
}
\end{table}

\textbf{Multiresolution \name is better for Unsupervised Registration.} We employed a multi-resolution architecture to enhance unsupervised registration results on the OASIS dataset. The outcomes, detailed in Table \ref{tab:MR-OASIS:Results}, demonstrate that incorporating multi-resolution features improves registration accuracy across all cases. Notably, the best performance is achieved with \name when using patch embedding blocks with strides of 2 and 4 (refer to Figure \ref{fig:architecture}B).

\textbf{ReMIND2Reg Results.} The \textit{ReMIND2Reg} dataset presents two key challenges: (a) It comprises multi-modal data, which introduces complexity in processing and analysis, and (b) It has a significantly smaller number of training samples—approximately half of those available in the OASIS dataset—further increasing the difficulty of achieving accurate results. Table \ref{tab:ReMIND2Reg:Results} presents the target registration error (TRE) and Jacobian determinant results on the ReMIND2Reg dataset. \name achieves the lowest TRE and the smallest percentage of negative values in Jacobian determinant of deformation among all methods. This indicates that \name not only excels in efficiency, with a much smaller model size, but also learns superior representations for multi-modal registration. Additionally, \name demonstrates robustness to limited dataset sizes, further widening the performance gap on smaller datasets. 

\begin{table}[!htb]
\caption{\textbf{ReMIND2Reg Unsupervised Registration Results.} Average Target Registration Error and Jacobian Determinant are obtained from Learn2Reg 2024 Challenge Page. \textit{Param} are listed in Millions of parameters used for training the model.} \label{tab:ReMIND2Reg:Results}
\centering
\small
\setlength{\tabcolsep}{1pt}
\scalebox{1.0}{

\begin{tabular}{c||c|c|c|c}
 \bf Methods & \bf C & \bf Param & \bf TRE $\downarrow$ & $|\mathbf{J}|< 0\%$ 
 $\downarrow$ \\
\hline
Siebert et al. \cite{siebert2021fast} & - & - & 3.87 $\pm$ 1.05 & 0.18 $\pm$ 0.009 \\
Fourier-Net\cite{jia2023fourier} & -& 1.1    &  4.128 $\pm$ 0.890 & 7.047 $\pm$ 1.113 \\
 \hline
 TransMorph-Tiny\cite{chen2022transmorph}&6 & 0.22  & 3.944 $\pm$ 0.693 & 0.013 $\pm$ 0.004 \\
 TransMorph\cite{chen2022transmorph} &96 &46.5  & 3.916 $\pm$ 0.77 & 0.024 $\pm$ 0.007 \\
 TransMorph-L \cite{chen2022transmorph} &128 & 108  & 3.902 $\pm$ 0.763 & 0.018 $\pm$ 0.003 \\
\hline

 EfficientMorph-11 &  96  & 1.8  &3.734 $\pm$ 0.798 & 0.011 $\pm$ 0.002 \\
 EfficientMorph-23 & 96  & 2.8  & \bf 3.599 $\pm$ 0.620 &  \bf0.010 $\pm$ 0.001\\ 
\end{tabular} 
}
\end{table}

\textbf{IXI Results.} Results of the IXI dataset are presented in Supplementary Table \ref{tab:IXI:Results}. 
\name outperforms traditional optimization-based methods such as SyN, NiftiReg, and various convolutional-based approaches such as VoxelMorph-H \cite{balakrishnan2019voxelmorph} and CycleMorph \cite{kim2021cyclemorph} by a significant margin. \name variants EM-11 and EM-23 with 4x4x4 strides achieve comparable performance (within $\pm$0.003) with less than 3 million parameters compared to TransMorph's 46 million parameters and 5$\times$ fewer epochs. Variants employing the Hi-Res tokenization technique with a stride 2 do not perform well for IXI. However, the ablations experiment with fewer embedding dimensions (C=24) improved the performance of 0.7317 to TransMorph's 0.7293 at 100 epochs, achieving similar accuracy as Fourier-Net-s. If trained for a longer period ($>100$ epochs), \name may probably be as accurate as TransMorph (maybe even higher), however this is left for future experiments. Accuracy vs epochs curves shown in supplementary Figure \ref{fig:IXI_curves} indicate that most \name variants outperform TransMorph in initial epochs, but then performance tends to saturate. Qualitative segmentations for the IXI dataset, shown in supplementary Figure \ref{fig:IXI_qualitative}, show that \name produces results of similar quality to TransMorph. For different substructures, EfficientMorph performs on par with the baseline, as shown in supplementary Figure \ref{fig:IXI_boxplot}.

\begin{table*}[!htb]
\caption{\textbf{Stride and Embedding Dimension Ablations.} Mean average dice score and standard deviation are evaluated on 35 segmented anatomies in OASIS. `stride' and `C' are the strides and embedding dimensions.} \label{tab:OASIS:Stride-and-Embedding-Dimension-Ablation:Results}
\centering
\setlength{\tabcolsep}{4pt} 
\scalebox{1.0}{
\begin{tabular}{c||c|c|c||c|c||c|c}
& && &\multicolumn{2}{c||}{\bf w/o Seg Loss} &  \multicolumn{2}{c}{\bf with Seg Loss} \\
\hline
 \bf Methods & \bf stride & \bf C & \bf Param(M)  &    \bf Dice Score $\uparrow$ &  $|\mathbf{J}|< 0\%$ $\downarrow$ & \bf Dice Score $\uparrow$ & $|\mathbf{J}|< 0\%$ $\downarrow$
 \\
\hline
 EfficientMorph-11 & 4x4x4 & 96 & 1.8 &0.795 $\pm$ 0.071 & 0.109 $\pm$ 0.012 &	0.841 $\pm$ 0.013 & 0.121 $\pm$ 0.017 \\
 EfficientMorph-23 & 4x4x4 & 96 & 2.8 & 0.796 $\pm$ 0.035 & 0.091 $\pm$ 0.0006  & 0.846 $\pm$ 0.013 & 0.125 $\pm$ 0.020 \\ 
 \hline
 EfficientMorph-11  & 2x2x2 & 96  & 1.8 & 0.803 $\pm$ 0.070 & 0.011 $\pm$ 0.002 & \bf 0.8623$\pm$0.0133 &  \bf 0.010 $\pm$ 0.001\\
 EfficientMorph-23 & 2x2x2 & 96  & 2.8  & \bf 0.810  $\pm$ 0.062 & \bf 0.010 $\pm$ 0.001 & \bf 0.870 $\pm$ 0.016 & 0.017 $\pm$ 0.001\\
\hline
 EfficientMorph-11  & 2x2x2 & 24  & 1.2  & 0.796 $\pm$ 0.067 & 0.108 $\pm$ 0.008 & 0.840 $\pm$ 0.011 & 0.125 $\pm$ 0.016 \\
 EfficientMorph-23 & 2x2x2 & 24  &  2.25	 & 0.799 $\pm$ 0.024 & 0.110 $\pm$ 0.014 & 0.8426 $\pm$ 0.013 & 0.126 $\pm$ 0.019\\
 \hline
  EfficientMorph-11  & 2x2x2 & 16  & 0.5  & 0.765 $\pm$ 0.004 & 0.164 $\pm$ 0.001 & 0.8311 $\pm$ 0.071 & 0.118 $\pm$ 0.011 \\
 EfficientMorph-23 & 2x2x2 & 16  &  1.3	 & 0.796 $\pm$ 0.003 &  0.149 $\pm$ 0.065 & 0.8345 $\pm$ 0.102 & 0.130 $\pm$ 0.102\\
\end{tabular} 
}
\end{table*}

\subsection{Ablation Studies}

Most ablation studies were conducted using the OASIS dataset. Additionally, specific studies, such as those on stride and embedding dimensions, were also carried out on the IXI dataset.

\begin{figure}[!htb]
    \centering
    \includegraphics[scale=0.30]{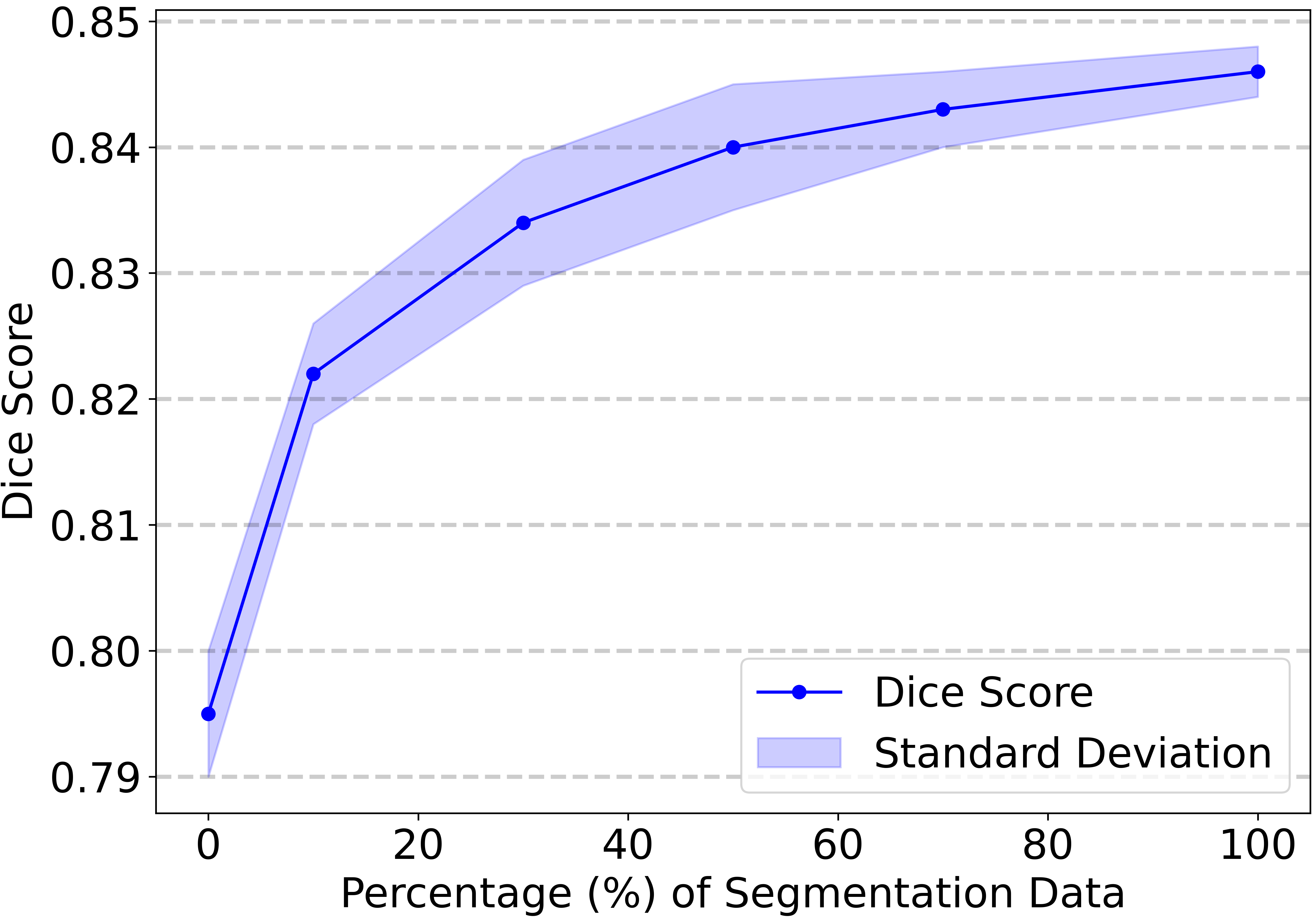}
    \caption{\textbf{Impact of Annotated Segmentation Available for Training.} These models were trained for EM-23 variant with stride 4x4x4 and embedding dimension 96.}
    \label{fig:seg_loss_percent}
\end{figure}

\textbf{Percentage of Segmentation Data.} Segmentation annotations are often unavailable for registration datasets, particularly in the medical field, where obtaining them is both time-consuming and labor-intensive. This challenge arises because multiple radiologists are typically required to mitigate human bias, which significantly increases the effort and time needed to generate accurate annotations. In this ablation study, we trained registration models with varying levels of segmentation data availability using the OASIS dataset. The results, shown in Figure \ref{fig:seg_loss_percent}, indicate that the performance curve is skewed. A substantial improvement in registration accuracy is observed when the initial 20\%-40\% of the dataset includes segmentations, but beyond this point, the relative performance improvements diminish with further increases in annotated data.

\textbf{Stride and Embedding Dimension Ablations.} We fully evaluate the impact of different hyper-parameters such as the stride of the voxel used for tokenization and the embedding dimension used in the patch embedding block. Results of these ablation studies are shown in Table \ref{tab:OASIS:Stride-and-Embedding-Dimension-Ablation:Results}. From the results, we see that increasing the embedding dimension with the same stride always performs better. Also, models trained with a smaller stride are always performing better, this proves that utilizing high-resolution spatial information for unsupervised registration results in better accuracy.

\textbf{Plane Order Ablations.} We also investigated the effect of varying the plane order (\textit{xy} vs \textit{yx}) in the EM-11 and EM-23 variants. The results of these experiments are shown in supplementary Table \ref{tab:OASIS:Plane-Variant-Ablation-EM-11:Results} and  Table \ref{tab:OASIS:Plane-Variant-Ablation-EM-23:Results}. The findings suggest that the order of plane attention has minimal impact on performance, as all variants cover all three volume axes, making the plane order unimportant.

\textbf{Attention Type Ablation.} We also explored the impact of various attention optimizations mentioned in related works, including Sparse \cite{child2019generating}, Linformer \cite{wang2020linformer}, Memory Efficient \cite{rabe2021self}, Nystrom \cite{xiong2021nystromformer}, and Flash \cite{dao2022flashattention}. The results, shown in supplementary Table \ref{tab:OASIS:Attention-Variant-Ablation-EM-25:Results}, indicate that for models using a stride of 4, different attention mechanisms have minimal effects on both performance and parameter count. This may be because these methods are optimized for processing billions of tokens, whereas 3D volumes typically involve only a few thousands tokens per sample. When experimenting with a stride of 2, we found that Flash attention reduced the parameter count by approximately 150k while maintaining similar performance to our best-performing EM-23 variant on OASIS dataset.

%% file: conclusion.tex
\section{Conclusion and Future Work}
We propose \name, a parameter-efficient transformer-based architecture for unsupervised 3D deformable image registration. 
\name uses a novel plane attention mechanism, which attends to 3D volumetric features by sequentially placing different plane attention blocks $xy$ followed by $yz$ or $zx$, thus attending to features along all three axes. Additionally, we propose a Hi-Res tokenization strategy to capture higher spatial resolution information while maintaining computational complexity. Evaluations of three datasets demonstrate that \name can achieve state-of-the-art results with a considerably lower parameter count ($\sim$16-27$\times$). \name with higher resolution token consumes larger memory while training, therefore in future work, we plan to explore memory-efficient model architectures using multi-resolution for 3D registration, segmentation, and synthesis applications. Additionally, incorporating frameworks such as Fourier-Net \cite{jia2023fourier} or SegFormer \cite{xie2021segformer} to reduce decoder complexity can further enhance the efficiency and effectiveness of our proposed model.

\section*{Acknowledgements}
\sloppy{The National Science Foundation supported this work under grant number NSF2217154. The content is solely the authors' responsibility and does not necessarily represent the official views of the National Science Foundation.}
\newpage

%% file: Supplementary.tex
\clearpage

\section{Supplementary}

\subsection{Additional Related Works}
The field of medical image registration is undergoing a paradigm shift with the rise of transformer-based methods, which are increasingly surpassing traditional convolutional neural networks (CNNs). While CNN-based approaches like SAME \cite{liu2021same} and SAMConvex \cite{li2023samconvex} relied on pre-trained models and carefully designed pipelines to capture global context, transformer architectures such as H-ViT \cite{ghahremani2024h} leverage multi-resolution high-level features to represent low-level voxel patches, overcoming the localized feature extraction limitations of CNNs.

Our proposed method advances medical image registration by introducing Hi-Res tokenization for efficient high-resolution feature extraction and a plane-based attention mechanism to balance localized focus with global context. Additionally, the multi-resolution variant captures spatial information early in the network, enabling the model to dynamically learn features across resolution scales and effectively capture intricate structural variations in medical images.

\subsection{Additional Dataset and Training Details}\label{supp:additional_details}

\subsubsection{Datasets Details} \label{supp:additional_dataset_details}
\paragraph{OASIS.} Preprocessing involves bias correction, skull stripping, alignment and cropping to dimensions of 160 x 192 x 224. Registration accuracy is reported by performing evaluation of corresponding segmentation masks for 35 anatomical structures. Additionally, FreeSurfer \cite{hoopes2021hypermorph} was used for pre-processing the brain MRI images, and it has label maps for 35 anatomical structures. Automated segmentation masks are used for evaluation of registration accuracy. 
\paragraph{Remind2Reg.} ReMIND2Reg dataset is a pre-processed subset of the ReMIND dataset \cite{juvekar2024remind}, containing multi-modal pre- and intra-operative data from patients who underwent brain tumor resection at Brigham and Women’s Hospital between 2018 and 2024. This dataset is part of the Learn2Reg 2024 challenge, which aims to register 3D iUS images with either ceT1 or T2 MRI images to account for brain shift during tumor resection, requiring models to handle large deformations and missing data scenarios. The dataset is divided into 155 image pairs for training, 10 image pairs for validation, and 40 for testing. The images are preprocessed into NIfTI format, cropped to a size of 256 $\times$ 256 $\times$ 256 with 0.5mm isotropic spacing, and co-registered where necessary. 
\paragraph{IXI.} The IXI dataset was augmented by flipping in random directions while training, as done by baselines. Evaluation was performed on corresponding segmentation masks for 29 anatomical structures with preprocessed size of 160 $\times$ 192 $\times$ 224.

\subsubsection{Further Training Pipeline Details of \name: Data Flow} \label{supp:additional_traiing_details}
Figure \ref{fig:architecture}A illustrates the proposed end-to-end registration network. The encoder processes two input volumes—a fixed and a moving image—by dividing them into non-overlapping 3D patches with dimensions $2 \times \frac{H}{S} \times \frac{W}{S} \times \frac{D}{S} \times C$, where $S={2,4,8}$. As $S$ increases, high-resolution features are progressively lost. For 
$S=2$, a Hi-Res Tokenization stage is employed to add depth-wise patch features and use a linear projection layer, enhancing fine-grained features while maintaining complexity similar to $S=4$ (see Section \ref{sec:appx:hi-res-tokenization}). This stage increases the channel dimension by d and incorporates positional encoding to preserve high-resolution feature tracking. For $S=4$ and $S=8$, Hi-Res Tokenization is omitted to see if Hi-Res block actually utilizing the features. 

Each patch, now termed as tokens, is processed through two efficient transformer blocks separated by a patch merging layer for downsampling. These transformer blocks may feature different plane attention modules ($xy \ \text{or} \ yz \ \text{or} \ zx$) depending on the variant. The bottleneck features from the encoder are then passed through convolutional layers (decoder) to generate a nonlinear 3D deformation field. This field is applied to the moving image using a PyTorch spatial transformer, similar to VoxelMorph \cite{balakrishnan2019voxelmorph} and TransMorph \cite{chen2022transmorph}.
The loss function integrates image similarity (local normalized cross-correlation) and regularization (bending energy) losses, following the approach of TransMorph \cite{chen2022transmorph} and Fourier-Net \cite{jia2023fourier}. The architecture utilizes default hyper-parameters weighted similarly to TransMorph’s \cite{chen2022transmorph}.

\subsubsection{Further Training Pipeline Details of Multi Resolution \name: Data Flow} \label{supp:additional_multires_details}
Figure \ref{fig:architecture}B illustrates the end-to-end data flow for the Multi-Resolution version of \name. The input is processed through two distinct patch embedding layers, each with different strides, $S_1^{'}$ and $S_2^{'}$. When a stride of 2 is used, the Hi-Resolution tokenization strategy described in Section \ref{sec:appx:hi-res-tokenization} is applied, allowing for efficient handling of high-resolution patches. Otherwise, the input flows through two parallel encoders with similar configurations, each capturing features at different resolutions.

The latent dimensions from these encoders are concatenated, enabling the model to leverage features across varying levels of detail. These bottleneck features are then passed into decoder to produce a nonlinear 3D deformation field, which is applied to the moving image via a PyTorch spatial transformer. The loss functions and hyperparameters remain same across both versions of the architecture, ensuring smooth integration of multi-resolution features while maintaining performance stability.

\subsection{Additional Ablation Experiments And Qualitative Results}
\begin{figure*}
    \centering
    \includegraphics[scale=0.28]{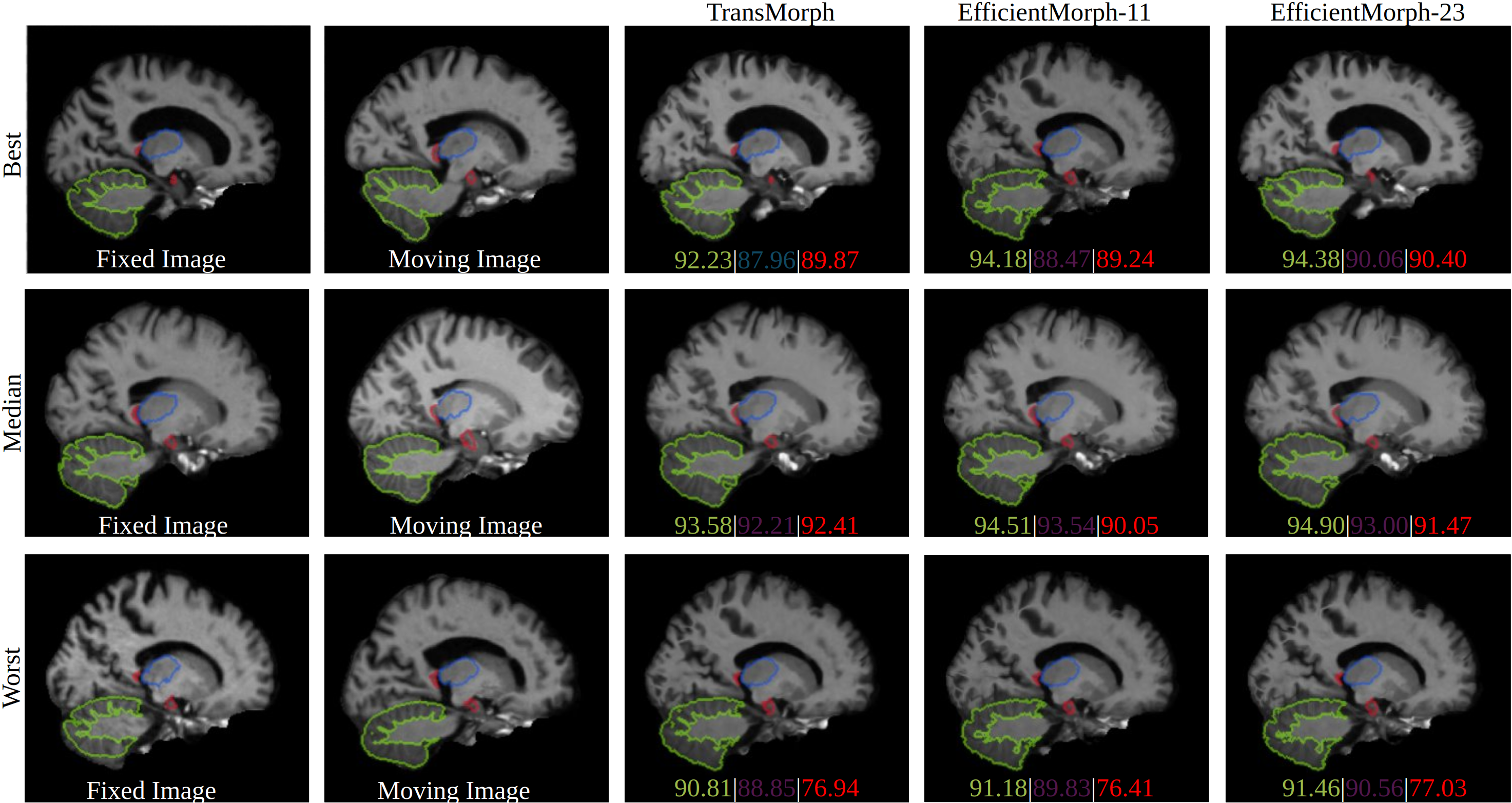}
    \caption{\textbf{OASIS qualitative results.} Comparison among the best, median, and worst output of TransMorph with the variants of the proposed method. Here, EfficientMorph-23 and EfficientMorph-11 are the different variants with 2x2x2 stride size and 96 embedded dimension; CGA means variants with cascaded group attention.}
    \label{fig:OASIS_qualitative}
\end{figure*}

\begin{figure*}
    \centering
    \includegraphics[scale=0.24]{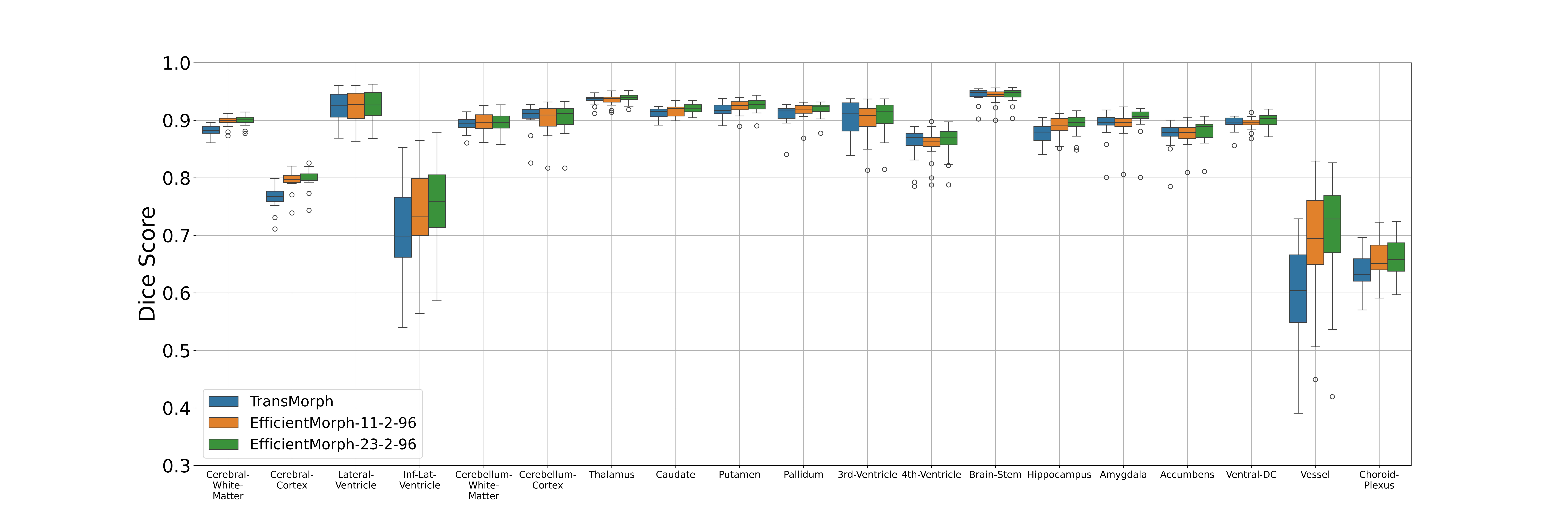}
    \caption{\textbf{OASIS boxplot.} Quantitative comparison of the proposed models with TransMorph showing dice scores for 19 anatomical substructures.}
    \label{fig:OASIS_boxplot}
\end{figure*}
\subsection{Qualitative Results}

Supplementary Figure \ref{fig:OASIS_qualitative} presents the best, median, and worst cases for both the baseline TransMorph \cite{chen2022transmorph} and the proposed model variants. It can be observed that the proposed variants achieve higher Dice scores for most anatomical structures, though performance slightly declines when segmenting smaller anatomies. Supplementary Figure \ref{fig:OASIS_boxplot} further compares the Dice scores of the proposed models with the baseline across 19 anatomical substructures, where the proposed variants consistently outperform the baseline in the majority of cases.

\subsection{Additional Ablation Results}

\begin{table}[!htb]
\caption{\textbf{Plane Attention Order Ablation.} Mean average dice score and standard deviation are evaluated on 35 segmented anatomies in OASIS with stride = 4 and C=96 for the EfficientMorph-11 variant.} \label{tab:OASIS:Plane-Variant-Ablation-EM-11:Results}
\centering
\setlength{\tabcolsep}{4pt} 
\scalebox{1.0}{
\begin{tabular}{c||c||c}
&{\bf w/o Seg Loss} & {\bf with Seg Loss} \\
\hline
 \bf Planes &    \bf Dice Score $\uparrow$ & \bf Dice Score $\uparrow$ \\
\hline

yz-xy & 0.795 $\pm$ 0.002 &	0.843 $\pm$ 0.0029  \\
xy-zx & 0.795 $\pm$ 0.005 & 0.844 $\pm$ 0.0033 \\ 

 yz-zx & 0.795 $\pm$ 0.003 & 0.844 $\pm$ 0.0032 \\
zx-xy & 0.795 $\pm$ 0.004 & 0.844 $\pm$ 0.0041 \\
zx-yz & 0.795 $\pm$ 0.003 & 0.843 $\pm$ 0.0036\\
\end{tabular} 
}
\vspace{-1em}
\end{table}

\begin{table}[!htb]

\caption{\textbf{Plane Attention - Pattern Ablations - EM-23.} Mean average dice score and standard deviation are evaluated on 35 segmented anatomies in OASIS with stride = 4 and C=96 for EM-23 variant with segmentation loss.} \label{tab:OASIS:Plane-Variant-Ablation-EM-23:Results}
\centering
\setlength{\tabcolsep}{4pt}
\scalebox{1.0}{
\begin{tabular}{c||c}

\bf Planes & \bf Dice Score $\uparrow$ \\
\hline
 xy-zx | zx-xy-yz  &	0.8378 $\pm$ 0.0040  \\
yz-xy |  zx-xy-yz & 0.8436 $\pm$ 0.0044 \\ 

yz-zx |  zx-xy-yz & \bf 0.8446 $\pm$ 0.0038 \\
zx-xy |  zx-xy-yz  & 0.846 $\pm$ 0.0042 \\

zx-yz |  zx-xy-yz & 0.843 $\pm$ 0.0037 \\
 zx-xy  | yz-xy-zx & 0.844 $\pm$ 0.0039 \\
zx-xy  | xy-yz-zx & 0.844 $\pm$ 0.0041 \\
\end{tabular} 
}
\end{table}

Supplementary Table \ref{tab:OASIS:Plane-Variant-Ablation-EM-23:Results} presents the results of the ablation study on plane order variants for the EfficientMorph-23 model. The table demonstrates that the proposed model achieves comparable accuracy regardless of the plane order.

\begin{table}[!htb]
\caption{\textbf{Attention Type Ablation.} Discuss about the different attention types that are added on top of the proposed plane attention, dice scores are evaluated on 35 segmented anatomies in OASIS with stride = 4 and C=96 for the EfficientMorph-23 variant.\textit{Param} as Parameter of model in millions of parameters.} \label{tab:OASIS:Attention-Variant-Ablation-EM-25:Results}
\centering
\setlength{\tabcolsep}{1pt} 
\scalebox{1.0}{
\begin{tabular}{c||c||c}
 \bf Attention & \bf Param & \bf Dice Score $\uparrow$ \\
\hline
 Plane & 2.8 & 	0.8458 $\pm$ 0.0137  \\
 Plane + Sparse \cite{child2019generating} & 2.8 & 0.843 $\pm$ 0.0040 \\
 Plane + Linformer \cite{wang2020linformer} & 4.69 &  0.848 $\pm$ 0.0035 \\

 Plane + Memory Efficient \cite{rabe2021self} & 2.82 &  0.848 $\pm$ 0.005 \\

 Plane + Nystrom \cite{xiong2021nystromformer} & 2.82 &  0.845 $\pm$ 0.0034\\

 Plane + Flash \cite{dao2022flashattention} & 2.82 & 0.845 $\pm$ 0.0042\\
 \hline
 Plane + Flash \cite{dao2022flashattention} (Stride$ = 2)$ & \bf 2.80& \bf 0.87 $\pm$ 0.0042\\
\end{tabular} 
}
\end{table}

Supplementary Table \ref{tab:OASIS:Attention-Variant-Ablation-EM-25:Results} highlights the evaluation of various memory-efficient attention mechanisms applied to the proposed plane attention. The results indicate that these approaches yield performance comparable to the proposed method, primarily because the tested mechanisms perform optimally with larger token sizes. Notably, Plane + Flash attention achieved the best performance when trained with a stride of 2, outperforming the stride 4 configuration supporting the claim.

\subsubsection{IXI dataset Results} \label{sec:appx:ixi_seg_loss}

Table \ref{tab:IXI:Results} presents the performance results on the IXI dataset. Notably, ablation experiments reducing the embedding dimensions (C=24) showed an improvement in performance from 0.7317, surpassing TransMorph's 0.7293 at 100 epochs. This also brought the model's accuracy in line with Fourier-Net-s while offering superior inference speed compared to all baselines. While extended training beyond 100 epochs could potentially result in even higher accuracy, this is left for future work.

\begin{figure*}[!thb]
   \centering
   \includegraphics[scale=0.23]{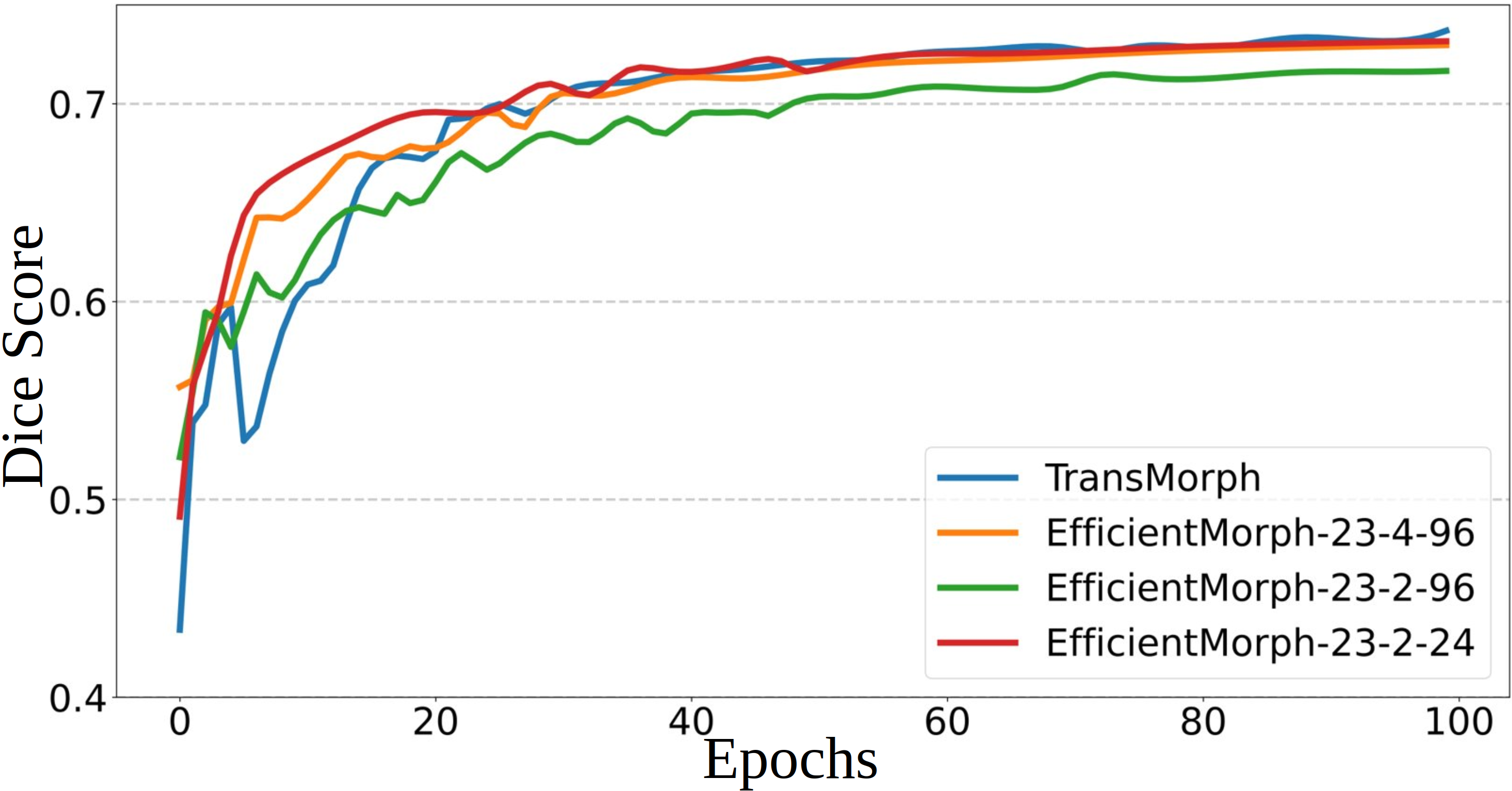}
   \caption{Dice scores as a function of number of epochs(IXI).}
   \label{fig:IXI_curves}
\end{figure*}

Supplementary Figure \ref{fig:IXI_curves} illustrates that EfficientMorph variants generally outperform TransMorph during the initial training stages, though performance plateaus as training progresses. Qualitative results in supplementary Figure \ref{fig:IXI_qualitative} indicate that EfficientMorph produces segmentation results comparable to TransMorph, with EfficientMorph performing similarly to the baseline for various substructures, as depicted in supplementary Figure \ref{fig:IXI_boxplot}.

\begin{table*}[!htb]
\centering
\caption{\textbf{IXI Results.} Mean average dice score and standard deviation are evaluated on 29 segmented anatomies in IXI. * indicates the performance numbers taken from TransMorph and Fourier-Net; for all others, we ran these baselines on our system for fair comparison. `stride' and `C' are the strides and channel layer for initial embedding layer. `Multi-Add' is the number of Multiply add operations needed for a forward pass.}
\label{tab:IXI:Results}
\small
\setlength{\tabcolsep}{8pt} 
\scalebox{1}{
\begin{tabular}{c|c|c|c|c|c|c}
& &  & & &\multicolumn{2}{c}{\bf Dice Score $\uparrow$} \\
\bf Methods & \bf stride & \bf C & \bf Epochs & \bf Param(M) & \bf Val & \bf Test \\
\hline
SyN*  & - & - & - & -  & - &0.645$\pm$0.152\\
NiftiReg* & - & - & -  & - & - &0.645$\pm$0.167\\
\hline
voxelMorph-1*\cite{balakrishnan2019voxelmorph} & -& - & - & 0.3 & - &0.548$\pm$0.317\\
cycleMorph*\cite{kim2021cyclemorph} & -& - & - & - &  - &0.528$\pm$0.321\\
Fourier-Net-s\cite{jia2023fourier}  & -& - & 200 &  1.05  &  0.729$\pm$0.024&0.730$\pm$0.025\\
Fourier-Net-s\cite{jia2023fourier}  & -& - & 1000 &  1.05 & 0.735$\pm$0.026&0.736$\pm$0.027\\
Fourier-Net\cite{jia2023fourier}*  & -& - & 1000 &  4.19  &  & \bf 0.760$\pm$0.132\\
\hline
TransMorph-Tiny*\cite{chen2022transmorph} & 4x4x4 & 6 & 500 & 0.24 &  0.545$\pm$0.180& 0.543$\pm$0.180\\
TransMorph\cite{chen2022transmorph} & 4x4x4 & 96 & 100 & 46.7  & 0.7293$\pm$0.029& 0.7324$\pm$0.0314\\
TransMorph\cite{chen2022transmorph} & 4x4x4 & 96 & 500 & 46.7 &  0.7405$\pm$0.0283&0.7408$\pm$0.0299\\
TransMorph-L\cite{chen2022transmorph}* & 4x4x4 & 128 & 500 & 108.34 &  0.753
$\pm$0.130& \bf 0.754$\pm$0.128\\
\hline
EfficientMorph-11 & 4x4x4 & 96 & 100 & 2.01  &  0.7233$\pm$0.0305 &	0.7224$\pm$0.0324\\
EfficientMorph-23 & 4x4x4 & 96 & 100 & 3.04  & 0.7291$\pm$0.0303 & 0.7298$\pm$0.0322\\ \hline

EfficientMorph-11 & 2x2x2 & 96 & 100 & 1.7 &	 0.6739$\pm$0.0322 & 0.6749$\pm$0.0323\\
EfficientMorph-23 & 2x2x2 & 96 & 100 & 2.8 &  0.7159$\pm$0.0307 & 0.7174$\pm$0.0330\\\hline
EfficientMorph-23 & 2x2x2 & 24 & 100 & 3.0 & \bf 0.7312$\pm$0.0298 & \bf 0.7317$\pm$0.0320 \\
\end{tabular}}
\end{table*}

\begin{figure*}[!thb]
    \centering
    \includegraphics[scale=0.41]{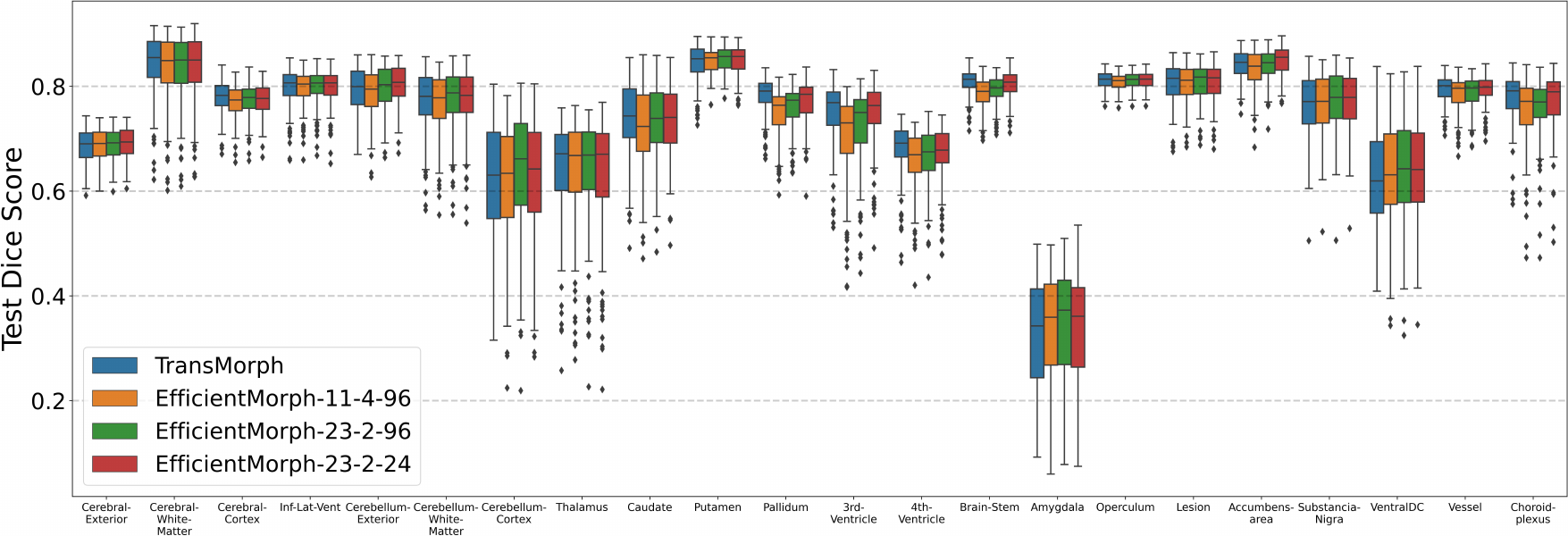}
    \caption{\textbf{IXI boxplot.} Quantitative comparison of the proposed models with TransMorph showing dice scores for 22 anatomical substructures.}
    \label{fig:IXI_boxplot}
\end{figure*}
\begin{figure*}[!thb]
    \centering
    \includegraphics[scale=0.22]{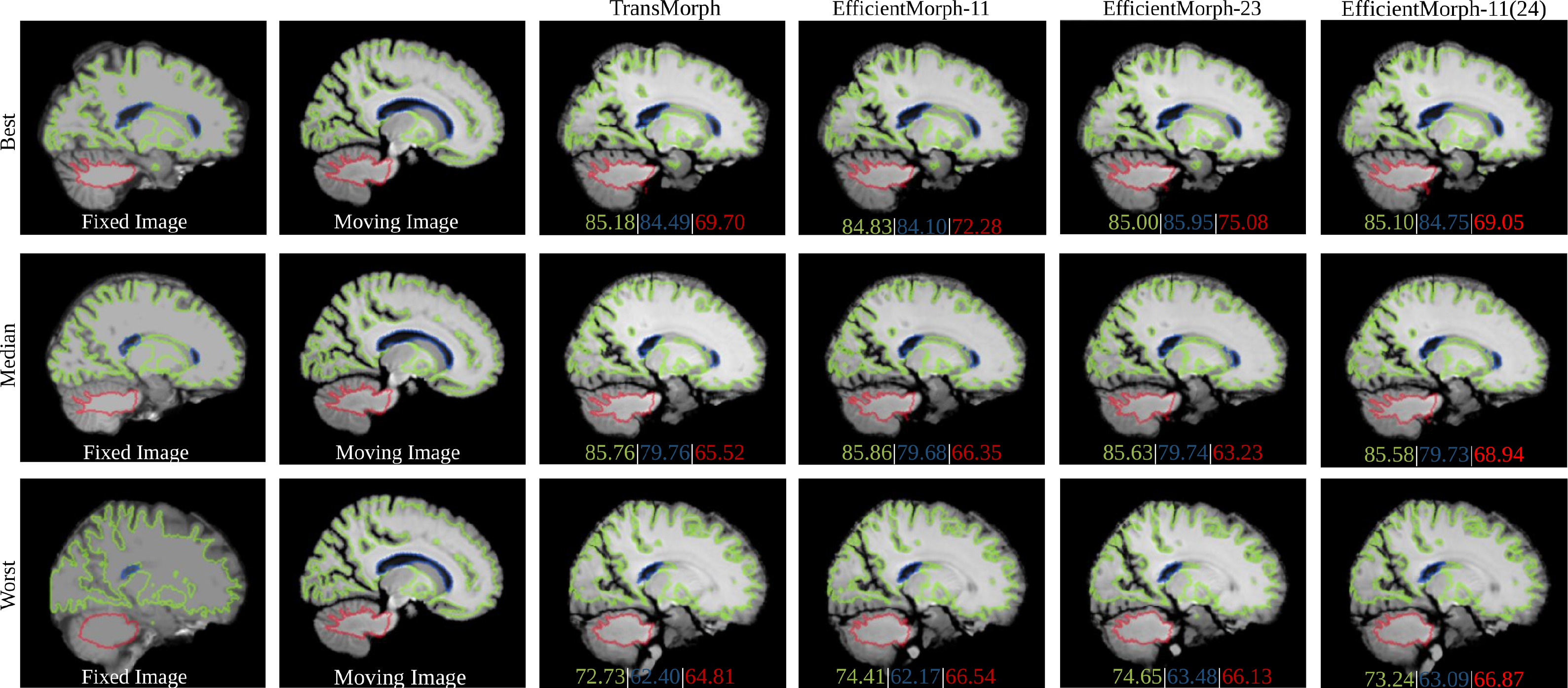}
    \caption{\textbf{IXI qualitative results.} Comparison among the best, median, and worst output of TransMorph with the variants of the proposed method. EfficientMorph-23 and EfficientMorph-11 are the different variants with 4x4x4 stride size and 96 embedded dimensions; EfficientMorph-11(24) has 24 embedding dimensions.}
    \label{fig:IXI_qualitative}
\end{figure*}